\documentclass[12pt,a4paper]{article}

\usepackage{amsmath,amssymb,amsthm,graphicx,color,bm,soul}

 \usepackage{graphicx} 
  \usepackage{here} 
	\usepackage[dvipdfmx]{hyperref}

\usepackage[normalem]{ulem}
\usepackage{cite}
\usepackage{bbm}
\newcommand{\rev}[2]{#2}
\newcommand{\id}{{\rm id}}
\newcommand{\deq}[1]{ \begin{align}#1\end{align}}

\newcommand{\deqn}[1]{ \begin{align}#1\end{align}}
\newcommand{\deqed}[1]{\begin{equation}\begin{aligned}#1\end{aligned}\end{equation}}
\newcommand{\deql}[2]{\deqn{\label{eq:#1}{#2}}}
\renewcommand{\And}{\wedge}
\newcommand{\bx}{{\bf x}}

\newcommand{\bA}{{\bf A}}
\newcommand{\bS}{{\bf S}}
\newcommand{\R}{{\mathbbm R}}
\newcommand{\cD}{{\cal D}}
\newcommand{\cH}{{\cal H}}
\newcommand{\cI}{{\cal I}}
\newcommand{\cL}{{\cal L}}
\newcommand{\Tr}{{\rm Tr}}
\newcommand{\ep}{\epsilon}

\newcommand{\rh}{\rho}
\newcommand{\si}{\sigma}
\newcommand{\ta}{\tau}
\newcommand{\av}[1]{\langle #1 \rangle}
 \newcommand{\Eq}[1]{Eq.~(\ref{eq:#1})}
  \newcommand{\eq}[1]{(\ref{eq:#1})}

\begin{document}

\title{Quantum-like cognition and decision making in the light of quantum measurement theory}

\author{M. Fuyama$^{1}$, A. Khrennikov$^{2}$ and M. Ozawa$^{3,4,5}$}

\maketitle

$^{1}$College of Letters, Ritsumeikan University, 56-1 Toujiin Kitamachi Kita-ku, Kyoto-Shi, Kyoto, 603-8577, Japan\\
$^{2}$Center Math Modeling 
in Physics and Cognitive Sciences
Linnaeus University, V\"axj\"o,  Sweden\\
$^{3}$Center for Mathematical Science and Artificial Intelligence\\  
Academy of Emerging Sciences, Chubu University, Kasugai, 487-8501, Japan\\
$^{4}$Graduate School of Informatics, Nagoya University, Nagoya, 464-8601, Japan\\
$^{5}$RIKEN, Hirosawa, Wako, 351-0198, Japan

{\bf keywords}: cognition, decision making, quantum-like modeling, quantum measurement theory, question order and response replicability effects, QQ-equality, sharp repeatable  non-projective measurements 

\begin{abstract}
We characterize the class of quantum measurements that matches the applications of quantum theory to cognition (and decision making) - quantum-like modeling. Projective measurements describe the canonical measurements of  the basic observables  of quantum physics.  However, the combinations of the basic cognitive effects, such as the question order and response replicability effects,  cannot be described by projective measurements. We motivate the use of the special class of quantum measurements, namely 
{\it sharp repeatable  non-projective measurements} - ${\cal SR\bar{P}}. $ This class is practically unused in quantum physics. Thus, physics and  cognition explore different parts of quantum measurement theory. Quantum-like modeling isn't automatic borrowing of the quantum formalism. Exploring the class  ${\cal SR\bar{P}}$  highlights the role of {\it noncommutativity  of the state update maps generated by measurement back action.} 
Thus,  ``non-classicality'' in quantum physics as well as quantum-like modeling for cognition 
is based on two different types of noncommutativity, of operators (observables) and instruments (state update maps): 
{\it observable-noncommutativity}  vs. {\it state update-noncommutativity}. 
We speculate that distinguishing quantum-like properties of the cognitive effects are the expressions of the latter, or possibly both.
\end{abstract}

\section{Introduction}

Quantum information revolution that is often called the second quantum revolution stimulated not only the development of quantum technologies, but even applications of the formalism of quantum theory outside of physics, especially to studies on   
cognition, consciousness, and decision making - {\it quantum-like modeling.}\footnote{In this note we can't reflect even the basic publications in this area. A Google search generates 213,000,000 results for ``quantum-like modeling''. We just mention a few  monographs \cite{QL0,UB_KHR,Busemeyer,Haven,QL3,Bagarello,Open_KHR,Bagarello1} and review \cite{Pothos}.} Cognitive tasks, e.g., questions asked to humans, are realized as measurements and quantum measurement theory  plays the crucial role in such studies.  In fact, this theory is more complex than it is often presented in textbooks. Typically (starting from von Neumann \cite{VN}) quantum measurements are described  as projective ones.\footnote{  We highlight that the notion of a projective measurement means not only that the probability distribution of its outcomes is based on the  projection valued measure (spectral measure of a \rev{Hermitian}{ self-adjoint } operator), but also that the measurement back action updates the original state by the projection onto the corresponding eigensubspace - the von Neuman-L\"{u}ders {\it projection postulate}.} The class ${\cal P}$ of projective measurements  covers the measurements of the  basic  quantum physical observables.\footnote{\rev{}{ Projective measurements, or more generally, accurate repeatable measurements are possible only for observables with discrete spectrum in the standard formulation of quantum mechanics \cite{O84,O85a}, while they are possible within arbitrarily small errors \cite{VN,O93}.}} The class ${\cal P}$ is also important for applications to cognition and decision making. The first stage of the development of quantum-like modeling was characterized by employment of the projective measurements \cite{Wang,Wang1}. However, later on it becomes clear that the class ${\cal P}$ doesn't cover all cognitive effects \cite{PLOS}.
We note that, although ${\cal P}$ is widely explored in quantum mechanics, this is only a special class of quantum measurements. 
	
This is a good place to make the following terminological remark. 
We  distinguish the notions  ``measurement'' and ``observable''.

``Measurement'' is characterized by a pair of its statistical properties: the ``outcome probability map'' 
describing the probability distribution of its outcome for any input state and the  ``state update map'' describing the state change resulting from measurement. 
Here, we only consider real valued measurements, the outcomes of which are represented as real numbers. 
In quantum theory, the outcome probability map is represented by a  positive (or probability)
operator valued measure (POVM). 
Thus any measurement is represented by two components, the POVM and 
the state update map. In quantum theory they are unified in the sole notion of  ``quantum instrument'' \cite{Davies-Lewis,DV,O84,O85a,O85b,O86,O97,O04,O23}.
Thus, we shall use the terms quantum measurement and instrument interchangeably.\footnote{ The tricky point is that in the projective measurement the state update is described by the same spectral projection valued measure as the probability distribution.
One might miss that this projection valued measure carries two totally different meanings. To be rigorous, one should speak about 
projective instruments, not just projective observables.}  

\rev{}{
To see this briefly, let ${\cal H}$ be the complex Hilbert space describing a quantum system $\bS$; for simplicity we work in finite dimensional spaces. 
Any state of the system $\bS$ is represented by a density operator $\rh$.
The space of linear operators on $\cH$ is denoted by ${\cal L}({\cal H}), $ the space of density operators
(i.e., positive operators with unit trace)  by $\cD(\cH)$.
A linear operator acting in ${\cal L}({\cal H})$ is called a {\em superoperator}. 
Consider a measurement, or more concretely, a {\em measuring apparatus} $\bA(\bx)$ with an {\em output variable} $\bx$ taking values $x$ in the real line $\R$.  
Let the ``outcome probability map'' be defined as 
\deq{\rh\mapsto\Pr\{\bx=x\|\rh\}}  
that maps the initial state $\rh$ (the state just before the measurement) to the probability $\Pr\{\bx=x\|\rh\}$ of obtaining the outcome $\bx=x$ in the initial state $\rh$.
Similarly, let the ``state update map'' be
\deq{
\rh\mapsto\rh_{\{\bx=x\}}
}
 that maps the initial state $\rh$ to the
updated state (the state just after the measurement) $\rh_{\{\bx=x\}}$ given the outcome $\bx=x$.  
The ``instrument'' $\cI$ associated with the apparatus $\bA(\bx)$ is defined by 
\deq{\label{eq:instrument}
\cI(x)\rh= \Pr\{\bx=x\|\rh\}\rh_{\{\bx=x\}}
}
for any $\rh\in\cD(\cH)$.
Then, as $\Tr[\rh_{\{\bx=x\}}]=1$, we obtain
\deq{
\Pr\{\bx=x\|\rh\}&=\Tr[\cI(x)\rh],\\
\rh_{\{\bx=x\}}&=\frac{\cI(x)\rh}{\Tr[\cI(x)\rh]}.
}
Thus, the instrument $\cI$ integrates both the ``outcome probability map'' 
$\rh\mapsto\Pr\{\bx=x\|\rh\}$ 
and the ``state update map''
$\rh\mapsto\Pr\{\bx=x\|\rh\}$ 
in a single mathematical object.
Now, It can be shown that 
the mapping $\rh\mapsto \cI(x)\rh$ defined in \Eq{instrument}
is naturally extended to a positive superoperator 
on $\cL(\cH)$ for every $x\in X$,
where $X=\{x_1,...,x_n\}\subseteq \mathbb{R}$ is a set of possible outcomes.
 such that $\sum_{x\in X}\cI(x)$ is trace-preserving,
and these properties mathematically 
characterizes the notion of instruments
as originally introduced by Davies and Lewis \cite{Davies-Lewis}; see \cite{O23} for detail.}

In contrast,  ``Observable'' generally means a physical quantity that can be measured \rev{}{or observed},
where a physical property is included as a $\{0,1\}$-valued quantity.
In quantum formalism, observables of the system is represented by self-adjoint operators 
on the ``state space'' (the Hilbert space of state vectors) of the system,
and every observable takes its value as one of its eigenvalues in any state in the corresponding
eigensubspace.  
The projections onto the eigensubspaces is called the ``spectral measure of the observable'',
which is a projection valued (or projective) POVM, and conversely every projective POVM is associated to a unique observable in this way, \rev{}{according to the spectral theory for  self-adjoint operators. 
The probability of the outcome from the measurement of an observable in a given state 
is determined by the spectral measure of the observable using the Born formula. 
On the other hand, non-observable physical quantity has no corresponding self-adjoint operator. 
For example, the one dimensional quantum harmonic oscillator has the Hilbert space 
$L^{2}(\R)$
on which there are self-adjoint operators corresponding to position, momentum, and energy, 
but there is no self-adjoint operator corresponding to phase. Thus, for the one dimensional quantum harmonic oscillator, position, momentum, and energy are observables but phase is not an observable \cite{BV07}. }

{``Measurement'' and ``observables'' are coupled here. 
A measurement is called a ``measurement of an observable'' if its POVM is identical with the spectral
measure of the observable.
Thus, different measurements with the same projective POVM is considered to describe different ways 
(with different state update maps) of measuring the same observable corresponding to that POVM. 
One of such measurements is the ``projective measurement'' of that observable, 
the state update map of which projects the input state onto the eigensubspace corresponding to the outcome. }

Any measurement with non-projective POVM is considered as a measurement of some observable
with some error.
The uncertainty principle applies to any measurement to give a quantitative relation between 
the error of measuring one observable versus the disturbance in another observable; 
the universally valid form of the error-disturbance relation has been studied extensively 
for the last two decades \cite{O03,O04a,O04,ESSBOY12,Bra13,O19,SDOH21}.

\begin{figure}[H]
\begin{center} 
\includegraphics[width=1\textwidth]{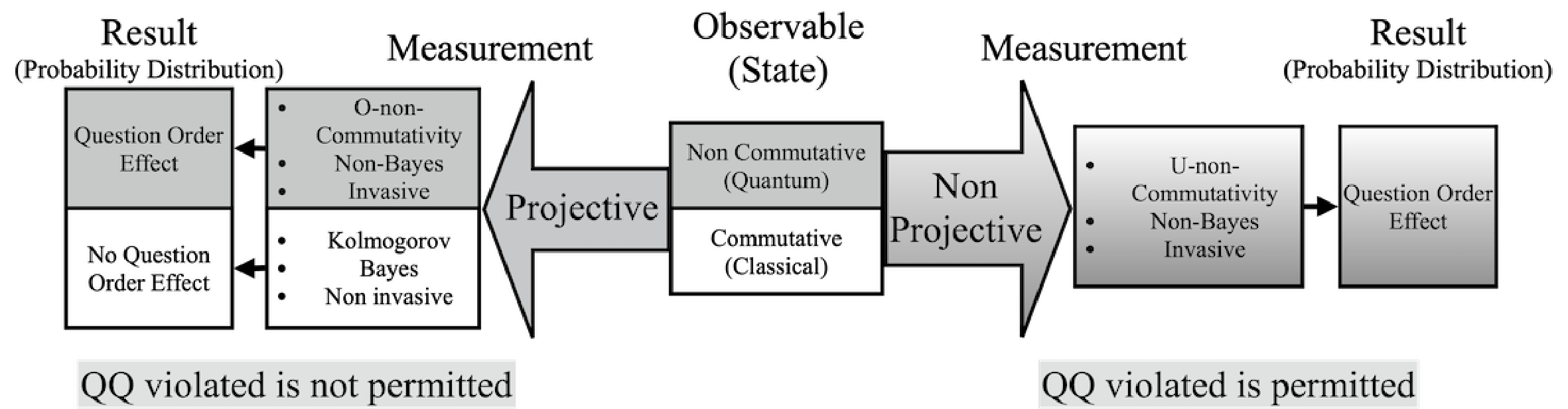}\hspace{8mm} 
\end{center}
\label{Masanao}
\caption{Schematic presentation of the structure of measurement theory, classical vs. quantum:
(see subsection \ref{se:caption} for detailed explanations and decoding of its blocks).}
\end{figure}

\rev{Besides of}{ In addition to } the one-to-one correspondence between the observables and
the projective POVM, general POVMs are often considered as 
generalized observables, especially in quantum information theory.
Such generalized  observables can also be useful in quantum-like modeling.
However, structuring  the discussion in the form ``observables versus POVMs'' is  wrong strategy, since such discussion 
would not highlight  the structure of state update maps. And from our viewpoint, the essence of the difference between physical and cognitive measurements is precisely in this structure.  The situation is complicated and its complexity  will be illuminated in this paper, see the scheme at Fig \ref{Masanao}.

Our main message (to the quantum-like community as well as to quantum physicists  who may become interested in new applications) is that to model cognition one should use a special part of  quantum measurement theory - {\it sharp  repeatable  non-projective measurements}
(the meaning of each term will be explained in section \ref{Types}). Denote this class of measurements by the symbol ${\cal SR\bar{P}}.$ Such measurements are practically unused in quantum physics, although they are covered by the general quantum measurement theory that studies all the physically realizable measurements. It seems that {\it physics (as described by conventional textbooks) and cognition basically coupled to two different classes of quantum measurements,} namely, ${\cal P}$ and ${\cal SR\bar{P}}.$ (By ``basically'' we mean that their characteristic effects are described within these classes.)\footnote{
Note that in quantum physics, nonrepeatable measurements (the class ${\cal \bar{R}}$) is also practically used for some basic observables such as photon counting measurements of the photon number observable.}

\sloppy
This search for a proper class of measurements for the quantum-like framework for cognition and decision making,  
leads to ``reincarnation'' of the old debate  on the quantum-classical interplay.  It seems that going outside of purely physical applications leads to the novel view on this interplay.  The employment of the class ${\cal SR\bar{P}}$ highlights the role of noncommutativity 
of the state update maps generated by measurement back action, see (\ref{eq:IC}), formulated within quantum instrument theory. Thus, non-classical effects are related to two different types of noncommutativity, of operators (observables) and instruments (state update maps). 
So, we should distinguish two types of noncommutativity, {\it observable-noncommutativity } (${\cal O}$-noncommutativity) and state {\it update-noncommutativity}
(${\cal U}$-noncommutativity). 

\rev{}{The order effect in quantum measurements is expressed as the ${\cal U}$-noncommutativity.
In quantum mechanics, it is well-known that the ${\cal U}$-noncommutativity follows from
the ${\cal O}$-noncommutativity according to the uncertainty principle.
However, in quantum-like modeling of cognitive effects, 
the ${\cal U}$-noncommutativity without the ${\cal O}$-noncommutativity plays an important 
role in quantum-like modeling of cognition and decision-making. }
We speculate that distinguishing quantum-like cognitive effects are the expressions 
of ${\cal U}$-noncommutativity. This doesn't exclude that   ${\cal O}$-noncommutativity also plays the important (but not crucial!) role in quantum cognition.

\rev{}{In the Wang-Busemeyer model \cite{Wang}, the order effect follows from the ${\cal O}$-noncommutativity of the projections representing a pair of questions, whereas in the Ozawa-Khrennikov model \cite{OK21} the order effect does not follow from the ${\cal O}$-non-commutativity, since the pair of questions are represented by two commuting projections. In both models, the order effect is formulated as  ${\cal U}$-noncommutativity of 
the belief-update map. In Kolmogorov's classical probability theory, all observables (i.e.,
classical random variables) commute and the belief-update maps are Bayesian,  or equivalently non-invasive, 
so they do not show the ${\cal U}$-noncommutativity.
Thus, modeling the question order effect needs non-Bayesian, or non-invasive, belief-updates maps.}
   
Our suggestion to identify advanced cognitive measurements as belonging to class ${\cal SR\bar{P}}$ is based on quantum-like modeling of the combinations of a few cognitive effects within quantum measurement theory, see articles \cite{OK20,OK21,OK23} on 
combining the question order  and response replicability effects (section \ref{QOER}).
These cognitive effects involve sufficiently advanced cognition, using memory (at least short term) and the conscious state of mind. And our aim is to characterize such sort of cognitive measurements. Less advanced cognitive measurements, e.g., as in psychophysics,  need not belong to ${\cal SR\bar{P}}$ (nor ${\cal P}).$ 

In addition we point to the role of the QQ-equality (section \ref{QOER}, equality (\ref{BE})) in characterization of the class of 
advanced cognitive measurements. In article \cite{Wang} this equality was derived for ${\cal P}$-measurements and it was demonstrated 
that it holds for a plenty of data from social opinion polls  \cite{Wang1}. However, recently it was found that some data  
on aesthetic perception in literature shows the violation of the QQ-inequality \cite{Miho}. Hence, such data can't be described within the class
projective measurements ${\cal P}.$ At the same time 
measurements from the class  ${\cal SR\bar{P}}$ can both satisfy and violate this equality.

We briefly discuss the issue of (non-)invasiveness of measurements. Invasiveness is equivalent to the possibility to violate the formula 
of total probability (FTP) and to non-Bayesian update of probabilities. A consistent discussion on this issue is possible within 
von Neumann algebra framework - ``non-commutative probability theory'';  within it one can compare the quantum and classical measurement theories based respectively on the state spaces consisting of density operators and probability measures \cite{OClQ}.    
We just point out that quantum observables can be measured only invasively (except for trivial observables corresponding to constant quantities). 
We stress that the violation of FTP (and the employing non-Bayesian update) is the probabilistic expression of the disjunction effect \cite{UB_KHR,Busemeyer}.

Preparation of the concrete mental states  is a complex problem. Therefore in quantum-like modeling working within the state independent formalism isn't so natural. It may be useful to work with state dependent statements, say, instead of consideration of noncommuting operators, $[  A,   B] \not=0,$ to consider operators noncommunting in the concrete state $\rh$ (see  \cite{OK23}). 
We shortly discuss state dependent approach in section \ref{SDP}; we also mention state dependent non-invasiveness (validity of FTP for some density operator 
$\rh)$ in section \ref{Math}. 
    
The paper is of the conceptual nature and also contains a short review on the mathematical basis of quantum measurement theory (section \ref{Math}). The review is brief, but goes deeply in the advanced part of this theory, the calculus of quantum instruments.

\rev{}{
\subsection{Detailed explanations and decoding of the blocks of the schematic representation}
\label{se:caption}
At the center of the figure, the algebraic structure of the observable quantities (so-called observables, physical quantities, random variables) of the system to be measured is classified based on observable-non-commutativity or observable-commutativity.
Observable-non-commutativity leads to quantum probability theory,
while observable-commutativity leads to classical probability theory.
The left arrow from the central box indicates the role of projective measurements in both quantum and classical theories.
The upper section of the next box explains that in quantum theory,
projective measurements demonstrate ${\cal O}$-non-commutativity (i.e., the non-commutativity of state-update maps that is derived from  observable-non-commutativity), describe invasive measurements, and lead to non-Bayesian state updates (or belief updates, if the state represents a belief rather than the object's state).
In this context, the state update is also known as wave packet collapse in quantum mechanics.
The von Neumann-L\"{u}ders projection postulate in the conventional formulation of quantum mechanics requires that every measurement of an observable be a projective measurement. 
The lower section of the next box explains that in classical theory,
projective measurements describe non-invasive measurements 
and lead to the Bayesian state updates (belief updates).
The Kolmogorov probability theory postulates that every measurement of a random variable should be projective when 
defining the sequential joint probability distribution of arbitrary random variables, from which the Bayesian update rule (from the prior probability to the posterior probability) follows. 
The leftmost box explains that projective measurements of any pair of questions represented by non-commuting projections in quantum theory exhibit the question order effect, as used in the Wang-Busemeyer model \cite{Wang},
whereas projective measurements of any pair of questions in classical theory do not exhibit this effect.
The right arrow from the central box indicates the role of non-projective measurements in both quantum and classical theories.  
The next box explains that non-projective measurements demonstrate ${\cal U}$-non-commutativity 
(i.e., the non-commutativity of state-update maps not necessarily derived from observable-non-commutativity), 
describe invasive measurements, and lead to non-Bayesian state updates (belief updates).
The rightmost box explains that non-projective measurements of a pair of questions represented by projections, whether or not they are  
non-commuting, show the question order effect.
The question order effect caused by non-projective measurements of a pair of questions represented by commuting projections
is shown to be compatible with the response replicability effect in the Ozawa-Khrennikov model \cite{OK20,OK21}.
The leftmost bottom line explains that a pair of projective measurements of two projections (questions) shows the QQ-equality, while the rightmost bottom line explains that a pair of non-projective measurements does not necessarily show this equality.
}

\section{Mathematical formalism of quantum measurement theory}
\label{Math}


A projective measurement  is described 
by a {\it projection valued measure.} Let $X=\{x_1,...,x_n\}\subseteq \mathbb{R}$ be a set of possible outcomes and let   $x\to   E_A(x)$ be a map from $X$ to projections such that $  E_A(x)  E_A(y) = \delta_{x,y}  E_A(x)$ and 
$\sum_x   E_A(x)=I.$ This map determines the projection valued measure defined on the subsets $\Delta$ of $X$ as $\mu_A(\Delta)= \sum_{x \in \Delta}
E_A(x).$ For the quantum state given by the density operator $  \rh,$ the probability of the outcome $A=x$ is given by the Born rule 
$P(A=x|   \rh) = \rm{Tr}  \rh    E_A(x),
$
and the measurement with the outcome $A=x$ induces the state update 
\deq{
  \rh \mapsto    \rh_{x}  =\frac{  E_A(x)   \rh   E_A(x)}
 {\rm{Tr}[   E_A(x)  \rh    E_A(x)]}.
}
Since $  E_A(x)   E_A(x)=   E_A(x),$ this measurement is repeatable.  
{The self-adjoint operator $A$ representing the observable $A$ to be measured 
by this projective measurement is given by the spectral decomposition  $A=\sum_{j=1}^{n} x_j E_A(x_j)$;
the set $X$ is the spectrum (the set of eigenvalues) of $A$ and $E_A(x_j)$ is the projection operator 
onto the eigensubspace of $A$ corresponding to the eigenvalue $x_j$.}

Consider a measurement with outcome (variable) $A$ taking the discrete range of values 
$X=\{x_1,..., x_m,...\}.$
Any map $x \to {\cal I}_A(x),$ where for each $x\in X,$ the map  $ {\cal I}_A(x)$ is a positive superoperator and 
$
{\cal I}_A(X) = \sum_x {\cal I}_A(x): \cD(\cH) \to \cD(\cH),
$
or equivalently ${\cal I}_A(X)$ is trace-preserving, 
is called a {\it quantum instrument}\footnote{ This is the very general definition of quantum instrument given by Davies and Lewis 
\cite{Davies-Lewis,DV}. To make quantum instruments physically realizable, i.e., described on the basis of unitary interaction between 
a system $S$ and measurement apparatus $M,$ the class of quantum instruments should be restricted. The proper class - completely positive instruments - was found by Ozawa \cite{O84}.}{ with outcome $A$}.        
The probability for the outcome $A=x$ is given by the generalization of the Born rule  
\begin{equation}
\label{Obs6}
P_\rh(A =x) = \rm{Tr}\; [{\cal I}_A(x)   \rh].
\end{equation}
We note that a measurement with the outcome $A=x$ generates the state-update 
\begin{equation}
\label{Obs7}
  \rh \to   \rh_x= \frac{{\cal I}_A(x)  \rh}{\rm{Tr}[ {\cal I}_A(x)  \rh]}.
\end{equation}
Conditional probability $P(B=y|A=x\| \rh)$ is defined as
\begin{equation}
\label{CP}
P(B=y|A=x\| \rh) = P(B=y\|\rh_{A=x}) = \frac{\rm{Tr}[ {\cal I}_B(y) {\cal I}_A(x) \rh]}{\rm{Tr}[ {\cal I}_A(x) \rh ]}.
\end{equation}
By using the quantum conditional probability we define sequential join probability distribution,  
\begin{equation}
\label{JPD}
P(A=x, B=y\| \rh)=P(A=x\|\ \rh) P(B=y|A=x\| \rh)= \rm{Tr}[ {\cal I}_B(y) {\cal I}_A(x) \rh].
\end{equation}

Repeatability of measurement is described as $P(A=x|A=x\| \rh)= 1$ or 
$P(A=x,A=x\| \rh)= P(A=x\| \rh).$ This is $A-A$ replicability in RRE; $A-B-A$ and $B-A-B$ replicability are defined as 
\deqed{
\label{JPD1}
P(A=x,B=y,A=x\| \rh)&= P(A=x,B=y\| \rh), \\
 P(B=x,A=y,B=x\| \rh)&= P(B=x,A=y\| \rh)
}
It is crucial that the same observable $A$ can be measured  by a variety of instruments generating the same probability 
distribution, but different state updates. Each quantum instrument determines POVM  given by
$
  \Pi(x) = {\cal I}^{*}(x) I.  
$
Operators $  \Pi(x), x \in X,$ are called {\it effects}; they are positive semi-definite \rev{Hermitian}{ self-adjoint} and sum up to the unit operator:
$
\sum_{x \in X}   \Pi(x)=I.
$ 
An instrument is sharp if its POVM is of the projective type, i.e., all effects are mutually orthogonal projections. 
An instrument (measurement) belongs to the class ${\cal P}$ if its POVM is of the projective type and the same projections determine  the state update that is ${\cal I}_A(x)   \rh=    E_A(x)   \rh   E_A(x).$ An instrument belonging to the 
class ${\cal SR\bar{P}}$ has the projective POVM, but the state update map is not of the projective type. 

\subsection{Non-invasivenss vs. invasiveness}
Let ${\cal I}_A=({\cal I}_A(x))$ be quantum instrument with set of outcomes $X=\{x_1,...,x_m\}.$ ${\cal I}_A$ is called non-invasive 
if ${\cal I}_A(X)=\id$, where $\id$ is an identity superoperator.  Hence, for any density operator $\rh,$ 
$\sum_{x\in X} {\cal I}_A(x) \rh = \rh,$ 
or $\sum_{x\in X} p(A=x\|\rh)  \rh_{A=x} = \rh.$  Instrument ${\cal I}_A$ is called invasive if 
${\cal I}_A(X) \not= \id.$
 
Let ${\cal I}_A$ be a non-invasive instrument and let ${\cal I}_B$ be an arbitrary instrument.
Then 
\begin{equation}
\label{FTP}
\rm{Tr}[ {\cal I}_B(y) \rh]= \sum_{x\in X} p(A=x\|\rh)  \rm{Tr}[ {\cal I}_B(y)\rh_{A=x}] .
\end{equation} 
Thus, the {\it formula of total probability} (FTP) of classical probability theory holds 
\begin{equation}
\label{FTP1}
p(B=y\|\rh)= \sum_{x\in X} p(A=x\|\rh) p(B=y|A=x\| \rh)
\end{equation} 

The right framework for analysis of the interplay between non-invasive and invasive instruments is 
von Neumann algebra framework - ``non-commutative probability theory''.  In standard quantum formalism the notion of non-ivasiveness is trivialized by the following statement. A quantum instrument ${\cal I}_A$ is non-invasive if ${\cal I}_A(x)= k_x \id,$ where $k_x \geq 0, \sum_x k_x=1.$ Here it is natural to introduce the notion of the state dependent non-invasiveness. An instrument ${\cal I}_A$ is non-invasive in the state 
$\rh$ if ${\cal I}_A(X) \rh= \rh.$ This implies the validity of FTP (\ref{FTP1}) for any instrument ${\cal I}_B.$

\section{Disjunction, question order and response replicability effects, and QQ-equality}
\label{QOER}

In this section we briefly review some cognitive effects that might be considered as signatures of ``${\cal U}$-noncommutativity'' (see section 
\ref{Quantum} for the discussion of the meaning of this term). 

{\it Disjunction effect} (DE) is the violation of Savage Sure Thing Principle. Mathematically it is described as the violation of FTP (see (\ref{FTP1}) (see \cite{B0,UB_KHR,Busemeyer} for its quantum-like modeling).  

{\it Question order effect} (QOE) expresses dependence of the sequential joint probability distribution of answers to questions $A$ and $B$  on their  order that is $p_{AB} \not= p_{BA}$ (see (\ref{JPD}) for the definition of sequential join probability distribution).  This effect is well confirmed experimentally in cognitive psychology and sociology, e.g. \cite{Moore}. Its quantum-like modeling was \rev{preformed}{performed} by Wang and Busemeyer \cite{Wang} with projective measurements. 

{\it Response replicability effects} (RRE) is expressed as follows. Alice can be asked two questions $A$ and $B$ in different orders.  
Suppose that first she was asked the $A$-question and answered it, e.g.,  with  ``yes'': Then Alice is asked the $B$-question
and gives an answer to it. And then she is asked the $A$-question again; typically Alice would repeat with probability one her original answer to the $A$-question,  ``yes''. This effect is $A-B-A$
response replicability. In the absence of the intermediate $B$-question, this is $A-A$ replicability, or called 
{\em repeatability} in quantum theory. Combination of
$A-B-A$ and $B-A-B$ replicability forms RRE (see (\ref{JPD1}).

Why is it so important to consider the combination of QOE and RRE? Their coexistence show that QOE is present in the normal experimental condition, or to show that QOE is not a result of our irrationality or our lack of adequate memory. Under the condition that people can make consistent judgment using adequate memory, they show QOE.

We remark that in psychology and sociology DE and QOE were well studied, both theoretically and experimentally \cite{Moore}. In contrast, RRE have not yet been well studied. This effect attracted the attention of psychologists in connection with quantum-like modeling \cite{PLOS}.
 Up to now, 
only one experiment was performed \cite{BW}, while the interpretational issue was the subject of the hot debates (see comments to 
this article on PLOS One webpage; see also \cite{OKP}).The authors of that article claimed  that their experiment showed violation of RRE and the debate was concentrated on the validity of such interpretation for experiment's output. However, \rev{not}{ } the possibility \rev{}{ not }  to violate RRE was the seed of the interest to it within quantum-like modeling of cognition \cite{PLOS,OK20,OK21,OK23}. In contrast, the heuristically evident presence of this effect in decision making and its possible combination with QOE disturbs  the quantum-like project for cognition. As was shown in article \cite{PLOS}, combination QOE+RRE can't be modeled with ${\cal P}$-measurements. The experiment reported in article \cite{BW} can't help with the resolution of this problem.  Hence, one should recognize that the class ${\cal P}$  is not sufficiently wide to cover all cognitive effects and their combinations. And in \cite{OK20} it was suggested to leave this class and explore the measurements of the 
 class ${\cal SR\bar{P}}.$ 

As was found by Wang and Busemeyer a plenty of data collected in social opinion polls constrained by a special equality that they called {\it the QQ-equality} \cite{Wang}:
\deq{
\label{BE} 
q&=p(ByAy) +p(BnAn)- [p(AyBy) + p(AnBn)]\nonumber\\
&=p(AyBn) +p(AnBy)- [p(ByAn) + p(BnAy)] =0,
}
where $Ay, By$ and $An, Bn$ denote the answers ``yes'' and ``no'' to the quetions $A$ and $B.$ 
This equality can be derived  for quantum measurements belonging to the class ${\cal P}$ \cite{Wang}. 
In fact, this is one of  the tests 
for applicability of ${\cal P}$-measurements. {\it If experimental data doesn't satisfy the QQ-equality, then it can't be described 
with such measurements.}  We point out that such data was collected in the experiment reported in article \cite{Miho}. 

In \cite{Wang} this equality was discussed in connection with QOE. Quantities   
$q_y=p(ByAy)- p(AyBy)$ and $q_n=p(BnAn)-  p(AnBn)$ characterize the degree of QOE. Their equality can be treated as a special quantum-like constraint. In the Wang-Busemeyer model QOE was coupled to noncommutativity of the operators $  A$ and $  B$ representing the questions $A$ and $B$
 and hence incompatibility of these questions, where two observables or questions are said to be {\em incompatible} if
 the corresponding operators do not commute. 
\rev{}{However, in articles \cite{OK20,OK21} it was shown that QOE can be modeled by a pair of instruments measuring a pair of commuting observables represented by projection operators that have non-projective state update maps and at the same time satisfy the QQ-equality.}

\section{Sharp repeatable non-projective invasive measurements}
\label{Types} 

All quantum measurements are mathematically described as POVMs,
in the sense that every quantum measurement has a unique POVM that describes the
probability distribution of the outcome of the measurement. 
Typically measurements with projective POVMs 
 are referred to as sharp measurements and 
measurements with non-projective ones as unsharp. 

\rev{}{Incompatibility of observables $A$ and $B$ means that they cannot be measured simultaneously with complete accuracies for both observables. 
The Heisenberg uncertainty principle is typically treated as representing the trade off between the accuracies (or errors) of measurements of these observables. 
However, Heisenberg's original tradeoff relation \cite{Hei27} has been shown not to be universally valid and a universally valid reformulations has been first derived by Ozawa \cite{O03,O04}. 
Eventually, the violation of the original relation and validity of the new relation have been experimentally observed for certain pairs of observables \cite{ESSBOY12,RDMHSS12,
13EVR}.}

\rev{}{
To be more precise, for a pair of conjugate observables $Q$ and $P$, Heisenberg's original relation \cite{Hei27}
is of the form 
\deq{\ep(Q)\ep(P)\ge \hbar/2}
for their ``root-mean-square errors''\footnote{\rev{}{Heisenberg originally call $\ep(Q)$ ``die Genauigkeit, mit der der Wert $Q$ bekannt ist (the accuracy with which the value $Q$ is known)'' 
or  ``etwa der mittlere Fehler von $Q$ (approximately the mean error of $Q$)'' but did not give a precise
definition.  A precise definition of ``quantum root-mean-square error'' $\ep(A)$ for an arbitrary
measurement of an observable $A$ satisfying the requirements of soundness (if the measurement is accurate then $\ep(A)=0$) and completeness (if $\ep(A)=0$ then the measurement is accurate) is given in \cite{O19}.
A practical but not necessarily complete definition of $\ep(A)$ is given as the ``root-mean-square'' of error 
operator $M(\ta)-A(0)$, where in the Heisenberg picture for the measuring process, 
$A(0)$ is the measured observable $A$ just before the measurement, 
and $M(\ta)$ is the meter observable just after the measurement.
In contrast, the standard deviation $\si(A)$ in the initial state is the ``roor-mean-square'' of $A(0)-\av{A(0)}$,
which is independent of the measuring process and irrelevant to the measurement error or accuracy.
}}
$\ep(Q)$ and $\ep(P)$ with Planck's constant $\hbar$ divided by $2\pi$,
and Ozawa's relation is of the form 
\deq{\ep(Q)\ep(P)+\ep(Q)\si(P)+\si(Q)\ep(P)\ge\hbar/2}
with standard deviations $\si(Q),\si(P)$ in the initial state.
Kennard \cite{Ken27} derived another relation 
\deql{Kennard}{\si(Q)\si(P)\ge\hbar/2}
 for standard deviations
$\si(Q)$ and $\si(P)$, which is universally valid, 
but not relevant to the limitation to measurement accuracies in any way.
Those relations are extended to arbitrary pair of observables $A,B$ by replacing the lower bound $\hbar/2$
by $|\av{AB-BA}|/2$ introduced by Robertson \cite{Rob29} to generalize Kennard's relation \eq{Kennard}, 
where $ \av{\cdots}$ stands for the expectation value in the initial state.
}

We speculate that sharpness is one of the distinguishing features of advanced cognitive measurements, e.g. \rev{ questions such that answering to them }{ answering to a question }
 involves conscious information processing and leads to definite answers, say ``yes''-``no''. As was mentioned, the psychophysical tasks can be described by unsharp measurements.  Moreover, as was mentioned, data collected in  
aesthetic perception in literature  violates the QQ-equality \cite{Miho}. 
Such violation can be easily modeled with unsharp measurements \cite{Lebedev}. However, we want to emphasize that violation of QQ-equality can be modeled even with sharp measurements.

Sharpness and unsharpness only concern the POVMs of measurements. 
Now we turn to measurements
with their state update maps. 
(This is the good place to point out once again that 
the notion of POVM doesn't assume any concrete way of the state update.) 
 We now proceed to the next level of measurements' classification.   

The sharp measurements are classified as repeatable and non-repeatable. Denote the class of sharp repeatable measurements as ${\cal SR}.$  Here ``repeatability'' means that if the first $A$-measurement gives the outcome $A=x$ then 
the successive $A$-measurement (performed immediately after the first one) gives the same outcome $A=x$ with probability one.
(This is $A-A$ repeatability part of RRE.) 
 
We remark that unsharp POVMs are not relevant, since they do not show repeatability in the finite dimensional case \cite[Theorem 6.5]{O84}.
The latter is another motivation to describe advanced 
cognitive tasks as sharp observables, otherwise they can't be coupled to repeatable measurements. And repeatability is the important feature of advanced cognition characterized by employment of the short-term memory. Once again, in psychophysics 
repeatablity can be violated.  We note that projective measurements are repeatable:  ${\cal P} \subset {\cal SR}.$ But, ${\cal SR}\not= {\cal P},$ there exist {\it sharp repeatable non-projective measurements} and 
they are explored for application in cognitive psychology \cite{OK20,OK21,OK23}. The class of such instruments denoted as  ${\cal SR\bar{P}}$ seems to the proper class for modeling advanced cognitive measurements.   

Now we discuss the notions of {\it non-invasive} vs. {\it invasive} measurement. A measurement is non-invasive if, for any state $\rh,$ conditioning on measurement's outcomes  can be described by the classical formula of total probability - FTP, otherwise a measurement is invasive (see section \ref{Math} for details). We recall that violation of FTP can be interpreted as probabilistic interference 
\cite{QL0,UB_KHR,Busemeyer,Haven}.  Hence, invasive measurements can generate constructive and destructive interference of probabilities and the non-invasive can't. 
In applications to cognitive psychology the violation of FTP corresponds to DE. We remark again that the consistent analysis of the interplay between non-invasiveness and invasiveness can be performed within 
von Neumann algebra framework - ``non-commutative probability theory''.  We emphasize that all quantum measurements, besides trivial state updates, are invasive. 

Let $S$ be a cognitive system, e.g., a human.  We summarize our reasoning for exploring ${\cal SR\bar{P}}$-measurement:
\begin{itemize}
\item Sharp:  $S$ performs the accurate decisions of the ``yes''-``no'' type.    
\item Repeatable:  $S$ uses the short term memory to be able to repeat the answer to question $A$ immediately after answering to it.  
\item Non-projective: $S$'s memory  is stable to perturbations generated by answering to other questions (RRE).
\item Invasive: $S$ makes probabilistic judgments involving interference of probabilities, violating FTP and using non-Bayesian probability update (DE).   
\end{itemize}

We point out again that ${\cal SR\bar{P}}$ corresponds to advanced cognition, including conscious information processing.

\section{Quantumness: noncommutativity of observables versus state update maps}
\label{Quantum}

\subsection{${\cal P}$-class: noncommutativity $\sim$ noncommutativity of observables}

Since the work of Heisenberg \cite{Heisenberg} (written around 100 years ago) that led to employment of the matrix and later operator calculus \cite{VN} in quantum mechanics, {\it noncommutativity of the operators} representing observables has been considered as the main distinguishing feature of the mathematical formalism of quantum theory. Generally speaking, noncommutativity has been identified with the 
mathematical description of quantumness. This viewpoint is strongly supported by {\it the Heisenberg uncertainty relation}, e.g., in the form of the Schr\"odinger or more generally Robertson inequality. In turn, this relation led Bohr to the formulation of {\it  the complementarity principle} \cite{Bohr,PL2}, the basic methodological principle of quantum theory. 
Noncommutativity of operators is rigidly coupled to incompatibility of the corresponding observables that is the impossibility of their joint measurement or in other words observables $A$ and $B$ are jointly measurable if and only if the corresponding \rev{Hermitian}{ self-adjoint } operators $  A$ and $  B$ commute, $[  A,   B]=0.$ An algebra of mutually commuting observables is considered as a classical structure within quantum theory. Such observables can be jointly measured and they have the joint probability distribution determined by 
their joint projection valued measure. In this sense quantum physics can be considered as an extension of the classical physics. This is the good place to mention the Koopman-von Neumann representation of classical phase space mechanics in that the classical position and momentum variables are described as commuting \rev{Hermitian}{ self-adjoint} operators (cf., however, with \cite{Fritiof}).  This line of thinking on ``quantumness'' is common in physics (see also \cite{PLAG} for quantum-like theory).  

\subsection{ ${\cal SR\bar{P}}$-class: nonclassicality $\sim$ noncommutativity of state update maps} 

As was discussed, the applications of the quantum formalism outside of physics, e.g. to cognition and decision making, stimulate careful analysis of the interrelation of the  notions ``quantumness'' and ``classicality''. Modeling of the combination of QOE and RRE stimulated us to go outside of the class ${\cal P}$ and work with the measurements of the class   
$ {\cal SR\bar{P}}.$ The corresponding observables which used in this modeling \cite{OK20,OK21,OK23} commute! From the traditional viewpoint, such models should be treated as classical. However, reproduction of  QOE with $ {\cal SR\bar{P}}$-measurements prevents one from such conclusion.
Heuristically one understand that some sort of noncommutativity should be involved.  And this is noncommutativity of state update maps,
\begin{equation}
\label{eq:IC}
[{\cal I}_A(x), {\cal I}_B(y)]\not=0
\end{equation}
 for two quantum instruments ${\cal I}_A$ and ${\cal I}_B$ generating QOE. 
 	
 \rev{ This condition is necessary and sufficient for showing QOE. }{ In fact, 
  we say that {\em instruments $I_A$ and $I_B$ show the question order effect (QOE)},
  if 
  \deq{P(A=x,B=y\|\rh)\not=P(B=y,A=x\|\rh)}
  for some $x,y\rh$.
  Then, this is equivalent to the relation
  \deq{
  \Tr[[\cI_A(x),\cI_B(y)]\rh]\not=0.
 }
Thus, \Eq{IC} is a sufficient condition for $I_A$ and $I_B$ to show QOE.
}
 From this viewpoint, instruments belonging to the class $ {\cal SR\bar{P}}$ are nonclassical. At the same time the operators $  A$ and $  B$ representing the observables  in the quantum-like model for QOE+RRE must commute,
$[  A,   B]=0.$ The latter means that the observables can be jointly measurable, that they are compatible. And from this vewiwpoint, they should be treated as classical ones. 

Hence, there are two forms of non-classicality coupled to two forms of noncommutativity, for  observables and  state update maps. We can speak about {\it observables noncommutativity} (${\cal O}$-noncommutativity) and {\it update noncommutativity} 
(${\cal U}$-noncommutativity).  (We point out that for projective instruments, these two forms of noncommutativity are equivalent,
${\cal O}$-noncommutativity $\sim$ ${\cal U}$-noncommutativity.) As was highlighted, such classifications of noncommutativity was stimulated by development of quantum-like modeling. In physics the ${\cal P}$-class is plays the crucial role and, although generalized observables in the form of POVMs are actively used in quantum information theory, one typically doesn't understand that they are just derivatives of 
quantum instruments.\footnote{A recent paper \cite{O24} introduced a notion of observability
for generalized observables (or POVMs) called ``value reproducible measurability'' 
and proved that every sharp observables are value reproducibly measurable but every
unsharp generalized observables are not value reproducibly measurable. 
Thus, the observable status of unsharp generalized observables are questionable.}

It seems that the role of noncommutativity condition (\ref{eq:IC}) for quantum foundations was highlighted only recently \cite{OK20,OK21,OK23}. The order effect is not of high interest in physics. And the most important is that quantum phsyical systems don't exhibit RRE. Thus split of ``noncommutativity'' into two counterparts is the new foundational problem that came to scientists' attention through quantum-like studies.        
 
In the original quantum-like study (${\cal P}$-model) \cite{Wang,Wang1}, QOE was interpreted as the expression of noncommutativity of operators  $  A$ and $  B$ representing the questions $A$ and $B;$ that is  incompatibility of observables, the impossibility to determine their joint probability distribution. In this framework ``noncommutativity of cognition'' is interpreted as ${\cal O}$-noncommutativity, i.e., as in physics.  Our $ {\cal SR\bar{P}}$-model showed that generally such coupling to incompatibility of questions (tasks) 
is not supported by quantum measurement theory. Observables can have the joint probability distribution and, nevertheless, show QOE{, since the sequential joint probability distribution of the answers to
questions $A$ and $B$ differs from the joint probability distribution for the simultaneous
measurement of $A$ and $B$ due to the invasiveness of measurement; otherwise QOE does not present.

This situation has the following cognitive interpretation. At each moment a cognitive system $S$ (as a human being) performing advanced cognitive tasks involving memory (at least short term)  has the consistent probabilistic picture for possible answers to the questions $A$ and $B.$ One might say that the $ {\cal SR\bar{P}}$-model recovers {\it mental realism}. In less advanced processing of cognitive information $S$ can operate within ${\cal P}$-framework by demonstrating QOE and violating RRE.

\subsection{State dependence of noncommutativity and classicality}
\label{SDP}

\sloppy
As was emphasized in \cite{OK23}, the notions of noncommutativity and classicality are state dependent. 
First we look at  ${\cal O}$-noncommutativity.  
Consider observables  with \rev{Hermitian}{ self-adjoint} operators $  A_1,  A_2$ and spectral measures 
$  E_{A_i}(x), i=1,2.$   If for some state, e.g., a pure state 
$|\psi \rangle,$ \;  {$[ E_{ A_1}(x),   E_{A_2}(y)] |\psi\rangle =0$ for all $x,y$ } then the joint probability distribution is well defined,
{
\deq{\lefteqn{
p(A_1=x_1,A_2=x_2\|\psi)}\quad\nonumber\\ 
&= ||  E_{A_1}(x_1)   E_{A_2}(x_2) \psi||^2
 = ||  E_{A_2}(x_2)   E_{A_1}(x_1) \psi||^2\nonumber\\
& = ||  E_{A_1}(x_1) \And   E_{A_2}(x_2) \psi||^2,
}  
where $\And$ stands for the infimum of two projections.}
Thus, for the state $|\psi\rangle,$  $A_1$ and $A_2$  can be treated as classical observables. But, for another state
$|\phi \rangle$ such that  {$[ E_{A_1}(x), E_{A_2}(y)] |\phi\rangle \not=0$ for some $x,y$,  } observables   $A_1$ and $A_2$  behave as quantumly. 

{For the equivalence of state dependent commutativity and the existence of 
the joint probability distribution for a family $  A_1,...,  A_n$ of observables,
we have the following theorem} \cite[Therem 5.2]{O16}: A family $  A_1,...,  A_n$ of observables has its
joint probability distribution $p(A_1=x_1,\ldots,A_n=x_n\|\psi)$  in a state $|\phi\rangle$, i.e., 
 \deq{
p(A_1=x_1,\ldots,A_n=x_n\|\psi) &= ||  E_{A_1}(x_1)\cdots   E_{A_n}(x_n) \psi||^2\nonumber\\ 
&= ||  E_{A_1}(x_1) \And \cdots \And   E_{A_n}(x_n) \psi||^2.
}
if and only if 
$[f(A_1,...,  A_n), g(A_1,...,  A_n)]|\psi\rangle =0$ for any polynomials $f(A_1,...,  A_n)$
and $g(A_1,...,  A_n)$.

\bigskip

Consideration of such state dependent compatibility of observables is especially important in quantum-like modeling of cognition, where preparation of a cognitive system $S$ in an arbitrary state is practically impossible. Hence, state independent compatibility versus incompatibility is practically impossible to check.          

One can consider a weaker condition of state dependent commutativity \cite{OK23}
\begin{equation}
\label{SDC}
 \langle \psi | [  A_1,   A_2] |\psi\rangle =0 .
\end{equation}
It also implies certain classical features of observables (see \cite{OK23}). 

Now we turm to ${\cal U}$-noncommutativity. 
Consider two instruments ${\cal I}_1$ and ${\cal I}_2$ such that 
\begin{equation}
\label{SDC1}
[{\cal I}_1(x), {\cal I}_2(y)] \rh =0,
\end{equation}
for all pairs $(x,y).$ Then such instruments don't generate QOE for the state $\rh$ and from this viewpoint,
 they are classical in this state (even if the corresponding observables are projective and don't commute). 

\section{Concluding remarks}

This note contributes to the foundational analysis of quantum-like modeling, a variety of applications of quantum theory outside of physics, e.g., to cognition and decision making. The experience of the quantum-like studies shows that one can't simply borrow 
the quantum formalism and apply it to model cognitive effects. The lesson of the attempts to apply it to describe the combinations of a few basic effects in cognitive psychology is that the class of projective measurements ${\cal P}$ isn't wide enough for such purpose \cite{PLOS} (cf. \cite{Wang,Wang1}). At the same time this class covers the basic variables of quantum physics. Thus, it seems that physics and cognition are based on  different parts of quantum measurement theory. Our studies on combination QOE+RRE \cite{OK20,OK21,OK23} motivate to shape the quantum-like cognition within the class $ {\cal SR\bar{P}},$ sharp repeatable non-projective measurements. This class isn't 
{well } explored in quantum physics. Hence, quantum physics and quantum-like cognition (decision making) are basically coupled to different domains of quantum measurement theory.\footnote{   
This is the good place to repeat once again that the notion of measurement is mathematically formalized on the basis of quantum instrument calculus. An instrument unifies two objects, a POVM  that characterizes the probability distribution of outcomes and the state update map that characterizes the measurement back action. Hence, attempts to perform the foundational analysis of quantum-like modeling by using just the notion of observable or POVM is misleading; in particular, it is misleading to reduce this analysis to the interplay of  projective POVMs (observables) versus general POVMs. }We relate the class  $ {\cal SR\bar{P}}$ to advanced cognition involving memory; we speculate that human's consciousness operate with such measurements. Less advanced mental processing might be described by measurements that don't belong to  $ {\cal SR\bar{P}},$ in particular, by projective measurements. 

As was shown in \cite{OK21}, combination QOE+RRE and the QQ-equality can be portrayed within $ {\cal SR\bar{P}},$ but generally $ {\cal SR\bar{P}}$-measurements can violate the QQ-equality. The statistical data collected in the recent experiment 
on aesthetic perception in literature violates this equality \cite{Miho}. Hence, it can't be modeled within ${\cal P}.$
The QQ-equality (as well as RRE) can be used as the complimentary test to determine whether QOE can be portrayed with the projective measurements. The measurement basis for aesthetic perception should be analyzed in more detail. It isn't clear whether the class  $ {\cal SR\bar{P}}$  is proper for such perception. May be even the condition of sharpness is too strong. (We remark that instruments with unsharp  POVMs can violate the QQ-equality \cite{Lebedev}.)  

This paper is devoted to the analysis of  applicability of quantum measurement theory to cognition and decision making. Similar analysis is required for other areas of  quantum-like modeling, e.g., finances \cite{Accardi,HavenB,Haven1}.   

\section*{Acknowledgments} 

The authors were supported by  the JST, CREST Grant Number JPMJCR23P4; A.K. was partially supported by  the EU-grant CA21169 (DYNALIFE), 
and visiting professor fellowship to Tokyo University of Science (April 2024) by invitation of S. Iriyama. M.O. was partially supported by JSPS KAKENHI Grant Numbers JP24H01566, JP22K03424,
RIKEN TRIP Initiative (RIKEN Quantum), and the Quantinuum--Chubu University Collaboration in 2023--2024.

	\end{document}